\documentclass[a4paper, oneside, twocolumn, notitlepage, 10pt]{extarticle_ecoc}
\usepackage{ecoc}
\usepackage{soul}

\begin{document}
\selectlanguage{english}    


\title{Pre-, In-, and Post-Processing Class Imbalance Mitigation Techniques for Failure Detection in Optical Networks}



\author{
    Yousuf Moiz Ali\textsuperscript{(1)},
    Jaroslaw E. Prilepsky\textsuperscript{(1)},
    Nicola Sambo\textsuperscript{(2)},
    Jo{\~a}o Pedro\textsuperscript{(3)},
    Mohammad M. Hosseini\textsuperscript{(4)},\\
    Antonio Napoli\textsuperscript{(4)},
    Sergei K. Turitsyn\textsuperscript{(1)},
    Pedro Freire\textsuperscript{(1)}
}

\maketitle                  
\vspace{-3.5mm}

\vspace{-3.5mm}
\begin{strip}
    \begin{author_descr}

\textsuperscript{(1)}~Aston University, Birmingham, UK,
   \textcolor{blue}{\uline{y.moizali@aston.ac.uk}},
   \textsuperscript{(2)}~Scuola Superiore Sant'Anna, Pisa, Italy,
   \textsuperscript{(3)}~Nokia, Optical Networks, Carnaxide, Portugal, 
   \textsuperscript{(4)}~Nokia, Munich, Germany
   \end{author_descr}
\end{strip}

\renewcommand\footnotemark{}
\renewcommand\footnoterule{}


\begin{strip}
    \begin{ecoc_abstract}
                We compare pre-, in-, and post-processing techniques for class imbalance mitigation in optical network failure detection. Threshold Adjustment achieves the highest F1 gain (15.3\%), while Random Under-sampling (RUS) offers the fastest inference, highlighting a key performance-complexity trade-off. \textcopyright2025 The Author(s)
    \end{ecoc_abstract}
\end{strip}


\vspace{-6mm}
\section{Introduction}
\vspace{-1mm}
Machine Learning (ML) has gained considerable attention over the past years as one of the most promising tools in the management of failures in optical networks\cite{musumeci2019tutorial}. The introduction of ML has brought about its own set of challenges, such as a lack of good-quality datasets because network operators generally cannot share network data \cite{musumeci2019tutorial}. Another important issue is that even if a dataset is available, the distribution between normal and failure instances in the dataset is uneven, as normal (i.e., without failures) instances greatly outweigh the number of failure instances since optical networks are designed to be robust\cite{healy2025addressing}. Imbalanced training can lead to suboptimal performance; therefore, there is a clear need to find methods that efficiently tackle class imbalance and improve the performance of ML models.

This problem has already been studied in the literature. Pre-processing techniques (data-centric) such as data augmentation and generating synthetic data through generative AI (GenAI) techniques have been thoroughly investigated in the domain of failure detection and identification \cite{khan2022data, lun2023gan, khan2023data, xing2023failure, sambo2023potential, khan2024model, kruse2024monitoring, zhang2025shap, healy2025addressing}. In-processing approaches (model-centric), which directly modify the learning procedure of the ML algorithm, have also been investigated in the literature \cite{cichosz2021application, sun2023stacking, lun2023gan, khan2024model, zhang2025shap}. While these techniques improve the ML models, their effectiveness depends on the dataset used. 

\begin{figure*}[b]
    \centering
    \includegraphics[width=1.0\textwidth]{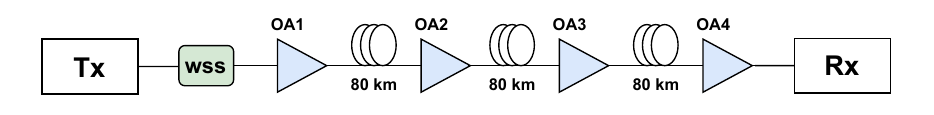}
    \caption{Experimental testbed setup}
    \label{fig:testbed}
\end{figure*}

To reduce the dependency on the quality of the dataset, post-processing or prediction-centric methods, which directly adjust the predictions from the ML model, can be very effective \cite{siddique2023survey}. To the best of our knowledge, they have not been explored so far in the area of class imbalance mitigation for failure detection. This paper presents the most comprehensive comparative study of pre-, in-, and post-processing techniques in terms of the number of methods tested for class imbalance mitigation in failure detection using an experimental dataset. The novelty of our approach is to find effective post-processing methods as an alternative or a complementary procedure to existing data-centric and model-centric techniques, which have been an untapped area in this domain. The pre-processing techniques explored include common sampling techniques such as SMOTE, Random Over-sampling (ROS), and GenAI techniques such as GANs and VAEs. The in-processing techniques include Bagging and Boosting, while post-processing methods include Threshold Adjustment and Cost-sensitive Thresholds. Our results indicate that post-processing approaches provide a higher F1 score compared to both pre-processing and in-processing techniques, with an improvement of up to 15\%.   

\vspace{-3mm}
\section{Class Imbalance Mitigation Techniques}
The class imbalance mitigation paradigm can be divided into three major categories: pre-processing, in-processing, and post-processing \cite{hort2024bias}. Pre-processing techniques modify the data before training, in-processing techniques alter the learning procedure of the model, and post-processing techniques modify the predictions from the trained ML model \cite{hort2024bias}.

\begin{figure*}[!t]
    \centering
    \includegraphics[scale=0.48]{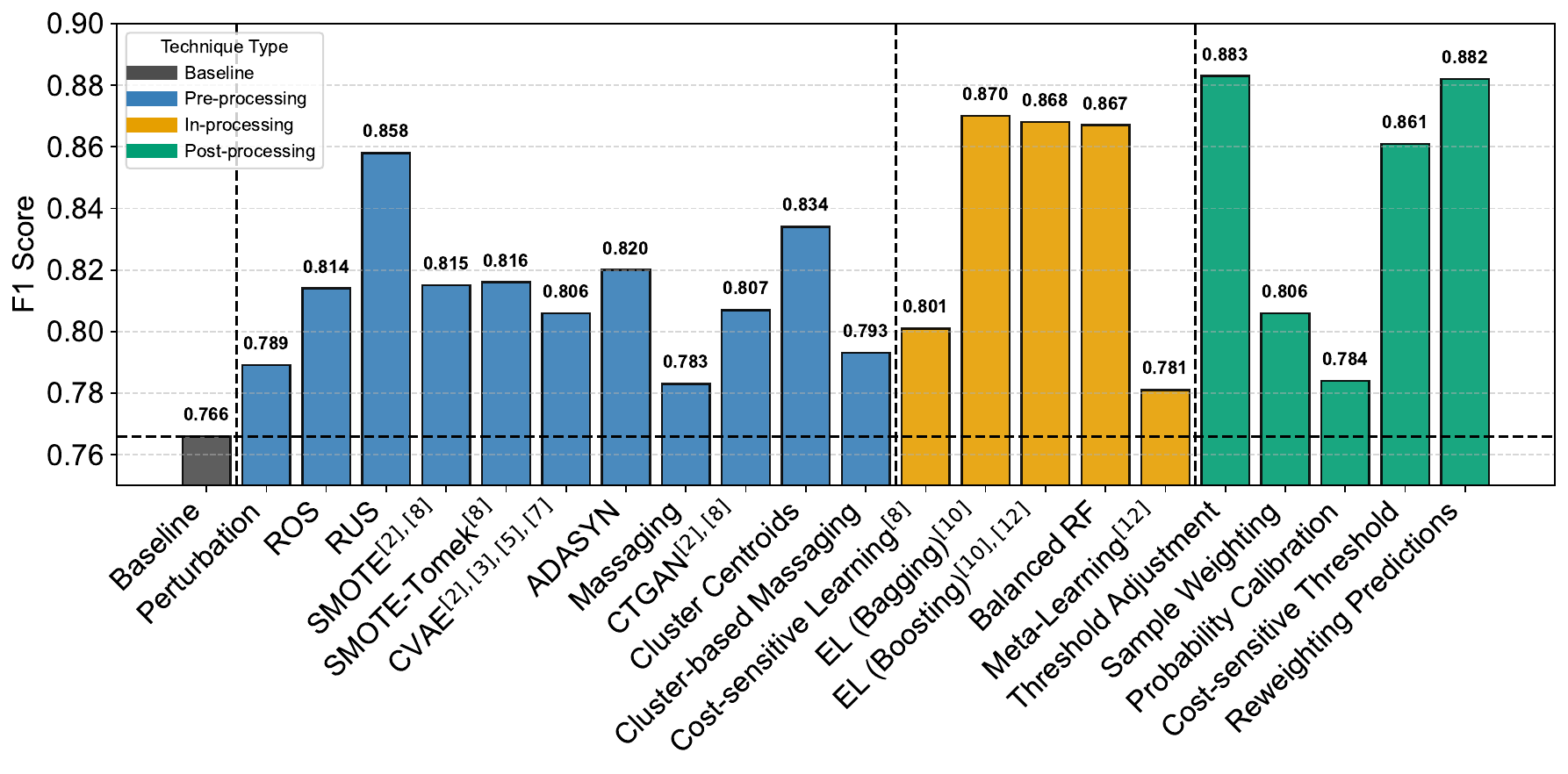}
    \vspace{-1.6mm}
    \caption{F1 score comparison of class imbalance mitigation techniques. Values in brackets indicate scientific papers that previously applied each technique for failure detection/identification in optical networks.}
    \label{fig:complete_results}
    \vspace{-2mm}
\end{figure*}

The pre-processing techniques tested in this study include over-sampling techniques such as ROS, SMOTE \cite{chawla2002smote}, and ADASYN\cite{he2008adasyn}, under-sampling techniques such as Random Under-sampling (RUS) and Cluster Centroids, and a combination of over-sampling and under-sampling technique such as SMOTE-Tomek\cite{batista2004study}. We also tested two GenAI techniques to generate synthetic samples: CTGAN\cite{ctgan} and CVAE\cite{ctgan}. The remaining techniques include Massaging\cite{hort2024bias}, Perturbation\cite{hort2024bias} and Cluster-based Massaging. 

In the in-processing category, we tested Cost-sensitive Learning, two Ensemble Learning (EL) techniques for Random Forests (RF): Bagging and Boosting\cite{liu2008exploratory}, Balanced RF (BRF)\cite{chen2004using}, and Meta-Learning, where we learn meta-features from a simple model before training an RF model on those meta-features.  

In the post-processing domain, the techniques applied include Threshold Adjustment, Cost-sensitive Threshold, Reweighting Predictions, Probability Calibration, and Sample Weighting.

\vspace{-3mm}
\section{Experimental Dataset and Baseline}
\label{sec:baseline}
\vspace{-0.5mm}
To test the techniques mentioned in the previous section, we used an experimental dataset generated in the labs at the Scuola Superiore Sant'Anna  \cite{silva2022learning}. Fig.~\ref{fig:testbed} shows the experimental testbed setup that comprises the transmitter (TX) and receiver (RX), a WSS to simulate failures, and a total of 3 fiber spans of 80 km with four optical amplifiers (OA). The features collected include the Timestamp, Type of device, ID of the device, BER and OSNR of the TX and RX, Input and Output powers of the OAs, and a binary Failure column\cite{silva2022learning}. For the sake of simplicity, we are considering an end-to-end monitoring system where we measure the BER and OSNR of the TX and RX. Originally, the data collected had 63248 normal samples and 2485 failure samples, which were further reduced to 7859 normal samples and 194 failure samples after doing some pre-processing and removing NaN values.

To establish a baseline for comparing class imbalance mitigation techniques, we selected the RF algorithm\cite{chen2004using}. RF was chosen due to its robustness in handling imbalanced datasets and relatively low computational complexity compared to more sophisticated models such as neural networks. The baseline results obtained using the original (imbalanced) dataset are presented in Tab.~\ref{tab:table1}.

For performance evaluation, we adopt the F1 score as the primary metric. Unlike accuracy, which can be misleading in imbalanced settings, the F1 score offers a more informative measure by accounting for both false positives and false negatives.

Each reported value in Tab.~\ref{tab:table1} represents the average over 100 independent runs to account for the stochastic nature of training and to provide a reliable estimate of performance variance. As shown, the baseline F1 score is relatively low, indicating poor generalization likely caused by the class imbalance. All subsequent experiments using mitigation techniques are evaluated relative to this baseline.
\vspace{-3mm}
\begin{table}[h!]
    \centering
    \caption{Baseline results on original dataset} \label{tab:table1}
    \begin{tabular}{|c|c|}
        \hline  \textbf{Metric} & \textbf{Score (average)}                 \\
        \hline  F1 Score & 0.7659                \\
        \hline
    \end{tabular}
\end{table}%

\vspace{-3.5mm}
\section{Results and Discussion}
\vspace{-0.5mm}
Fig.~\ref{fig:complete_results} presents the F1 scores achieved after applying various class imbalance mitigation techniques. Across all categories, we observe consistent improvements over the baseline established in the previous section. While other metrics such as accuracy, precision, and recall also showed gains, we focus on the F1 score here due to space constraints and its suitability for imbalanced classification tasks.

In the pre-processing category, RUS yielded the most significant improvement, increasing the F1 score by 12\% relative to the baseline. In contrast, techniques like Massaging and Perturbation demonstrated limited gains. These methods rely on strategic label flipping, which can distort the data distribution and lead to suboptimal generalization. Similarly, generative approaches showed marginal improvements, likely due to the limited separability in our dataset, which affects the quality of synthetic sample generation.

Among in-processing methods, the EL techniques and BRF outperformed others, delivering improvements of up to 13.6\%. These methods also surpassed the best-performing pre-processing technique (RUS) in F1 score. Cost-sensitive Learning and Meta-Learning approaches, however, showed only modest benefits.

In the post-processing category, Threshold Adjustment and Reweighting Predictions offered the most notable improvements, raising the F1 score by up to 15.3\% over the baseline. Cost-Sensitive Thresholding followed closely, while other methods contributed marginal gains. These findings indicate that post-processing techniques are the most effective for improving F1 performance. Additionally, their advantage lies in operating directly on model predictions, thereby reducing dependence on the underlying data quality—a common limitation of many pre-processing methods.

\begin{figure}[!t]
    \centering
    \includegraphics[scale=0.25]{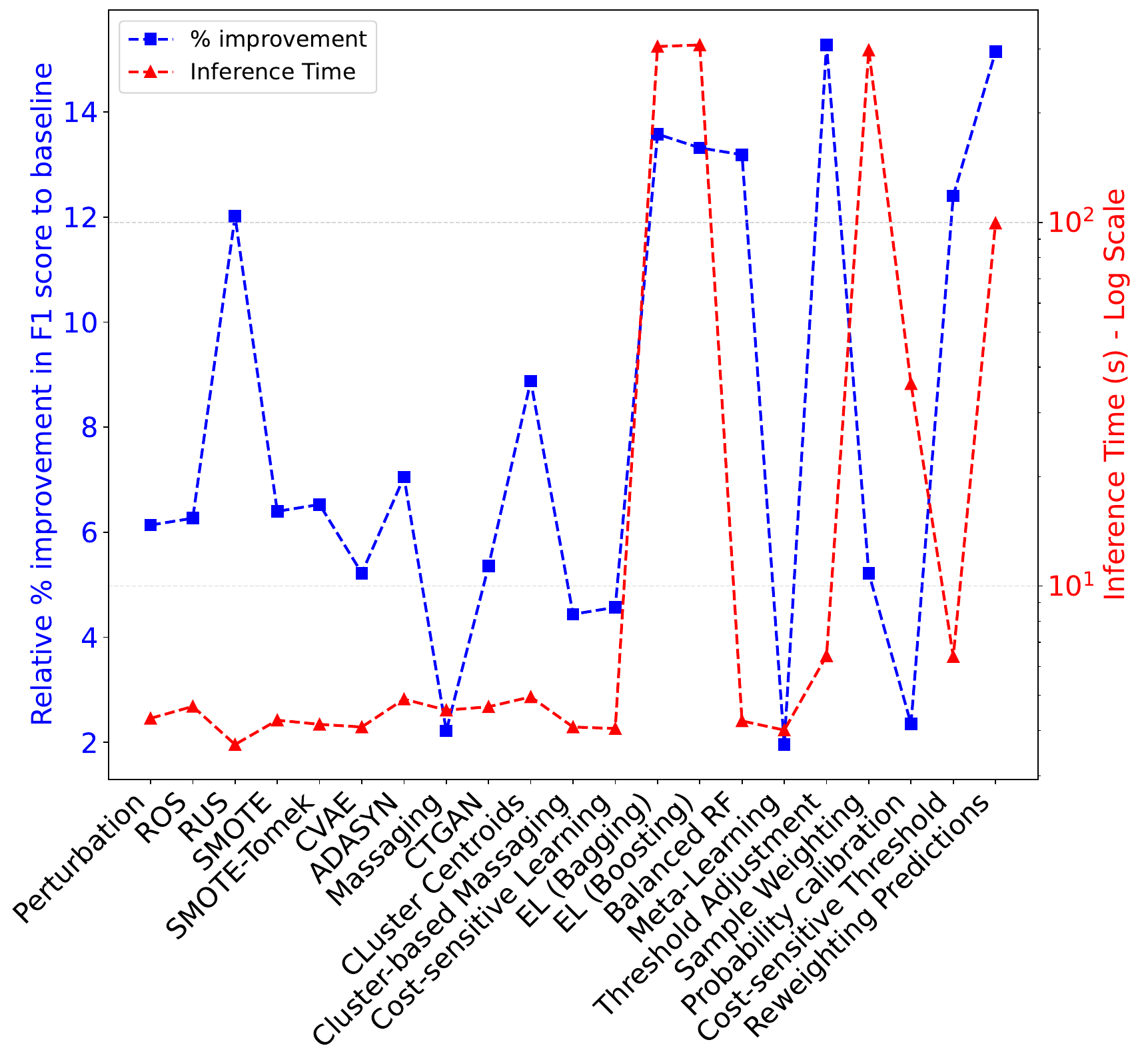}
    \caption{Percentage improvement in F1 score compared to the baseline with the inference time of the techniques.}
    \label{fig:percent_improv_run_time}
    \vspace{-2mm}
\end{figure}

Fig.~\ref{fig:percent_improv_run_time} presents a dual-axis plot illustrating the relative percentage improvement in F1 score over the baseline (left y-axis) alongside the corresponding inference times for each technique (right y-axis). An important observation from Fig.~\ref{fig:percent_improv_run_time} is that the inference times for all pre-processing methods and the majority of in-processing techniques remain comparable to the baseline. This is expected, as these methods either manipulate the training data or modify the learning process without introducing additional computational steps during inference. Fig.~\ref{fig:percent_improv_run_time} also illustrates the trade-off between inference time and performance improvement across the evaluated class imbalance mitigation techniques. As expected, the RUS technique exhibits the lowest inference time due to the reduced training dataset size, resulting in a simpler model. In contrast, the EL methods, which involve aggregating multiple RF models, incur the highest inference time owing to their increased computational complexity. \textit{The key insight from this result is the observed trade-off between model performance and inference efficiency.} Techniques positioned from left to right on the plot—from pre-processing to post-processing—generally show an upward trend in both inference time and percentage improvement in F1 score relative to the baseline. This trend underscores that achieving higher performance often comes at the expense of increased computational overhead.

\textit{ For applications prioritizing model performance}, Threshold Adjustment emerges as the most effective and relatively efficient method, offering the highest improvement in F1 score. Conversely, \textit{in latency-sensitive scenarios where inference speed is critical}, the RUS method is preferable due to its minimal computational burden while still delivering respectable performance gains.

\begin{figure}[htbp]
    \centering
    \includegraphics[scale=0.38]{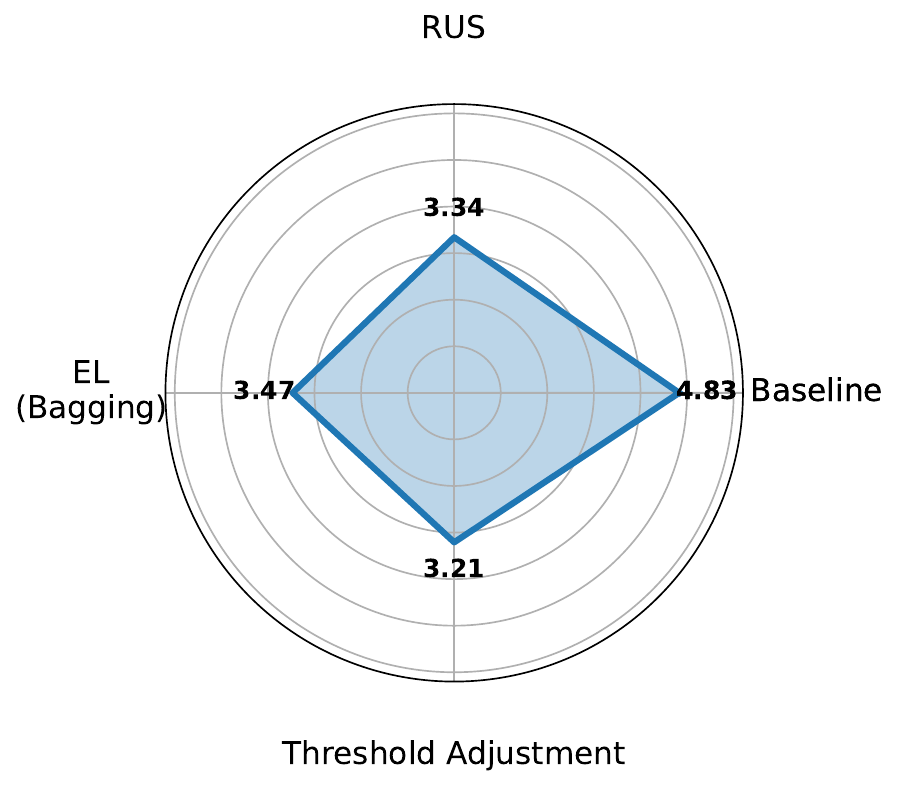}
    \caption{Variance-to-mean ratio for F1 score for the best techniques in each category and the baseline}
    \label{fig:variance_plot}
    \vspace{-1mm}
\end{figure}

Finally, Fig.~\ref{fig:variance_plot} presents the variance-to-mean ratio (VMR) of the F1 scores for the baseline and the top-performing technique from each category. The plot shows that (i) the VMR values for RUS, Bagging, and Threshold Adjustment are all lower than that of the baseline, and (ii) Threshold Adjustment achieves the lowest VMR overall. These results are noteworthy, as they demonstrate that beyond improving average F1 scores, these mitigation techniques also contribute to increased stability and consistency of model performance.

\vspace{-3.7mm}

\vspace{-1.4mm}
\section{Conclusions}

We studied the potential of post-processing class imbalance mitigation techniques for failure detection in optical networks, in addition to the pre-processing and in-processing methods. The results indicate that the Threshold Adjustment post-processing technique offers a more expressive improvement in the F1 score (15.3\% compared to the baseline). If inference time is critical, then the RUS technique may be a better-suited option. Finally, it has been shown that the best techniques from each category also improve the variance in the results compared to the baseline.

\clearpage
\section{Acknowledgements}
This work has received funding from the European Commission MSCA-DN NESTOR project (G.A. 101119983). SKT acknowledges EPSRC project TRANSNET (EP/R035342/1).


\defbibnote{myprenote}{%
}
\printbibliography[prenote=myprenote]

\vspace{-4mm}

\end{document}